\documentclass[runningheads]{502}
\usepackage{graphicx}
\usepackage{booktabs}
\usepackage{multirow}
\usepackage{xcolor}
\usepackage{cite}
\usepackage{float}

\begin{document}
\title{Incongruity Detection between Bangla News Headline and Body Content through Graph Neural Network}
%
%

\author{Md Aminul Haque Palash\inst{1}\orcidID{0000-0002-4794-8531} \and
Akib Khan\inst{2}\orcidID{0000-0002-1517-5319} \and
Kawsarul Islam\inst{1}\orcidID{0000-0003-2466-834X} \and MD Abdullah Al Nasim\inst{3}\orcidID{0000-0001-5594-607X} \and
Ryan Mohammad Bin Shahjahan\inst{3}}

 \authorrunning{Md et al.}
%

\institute{Department of Computer Science and Engineering, Chittagong University of Engineering and Technology,Chittagong 4349, Bangladesh, \and Brac University, Dhaka, Bangladesh, \and Pioneer Alpha, \email{info@pioneeralpha.com} }

%
%
\maketitle              
\begin{abstract}
 Incongruity between news headlines and the body content is a common method of deception used to attract readers. Profitable headlines pique readers' interest and encourage them to visit a specific website. This is usually done by adding an element of dishonesty, using enticements that do not precisely reflect the content being delivered. As a result, automatic detection of incongruent news between headline and body content using language analysis has gained the research community's attention. However, various solutions are primarily being developed for English to address this problem, leaving low-resource languages out of the picture. Bangla is ranked 7th among the top 100 most widely spoken languages, which motivates us to pay special attention to the Bangla language. Furthermore, Bangla has a more complex syntactic structure and fewer natural language processing resources, so it becomes challenging to perform NLP tasks like incongruity detection and stance detection. To tackle this problem, for the Bangla language, we offer a graph-based hierarchical dual encoder (BGHDE) model that learns the content similarity and contradiction between Bangla news headlines and content paragraphs effectively. The experimental results show that the proposed Bangla graph-based neural network model achieves above 90\% accuracy on various Bangla news datasets.

\end{abstract}
\section{Introduction}
\begin{figure}[H]
\centering
\includegraphics[width=0.5\linewidth]{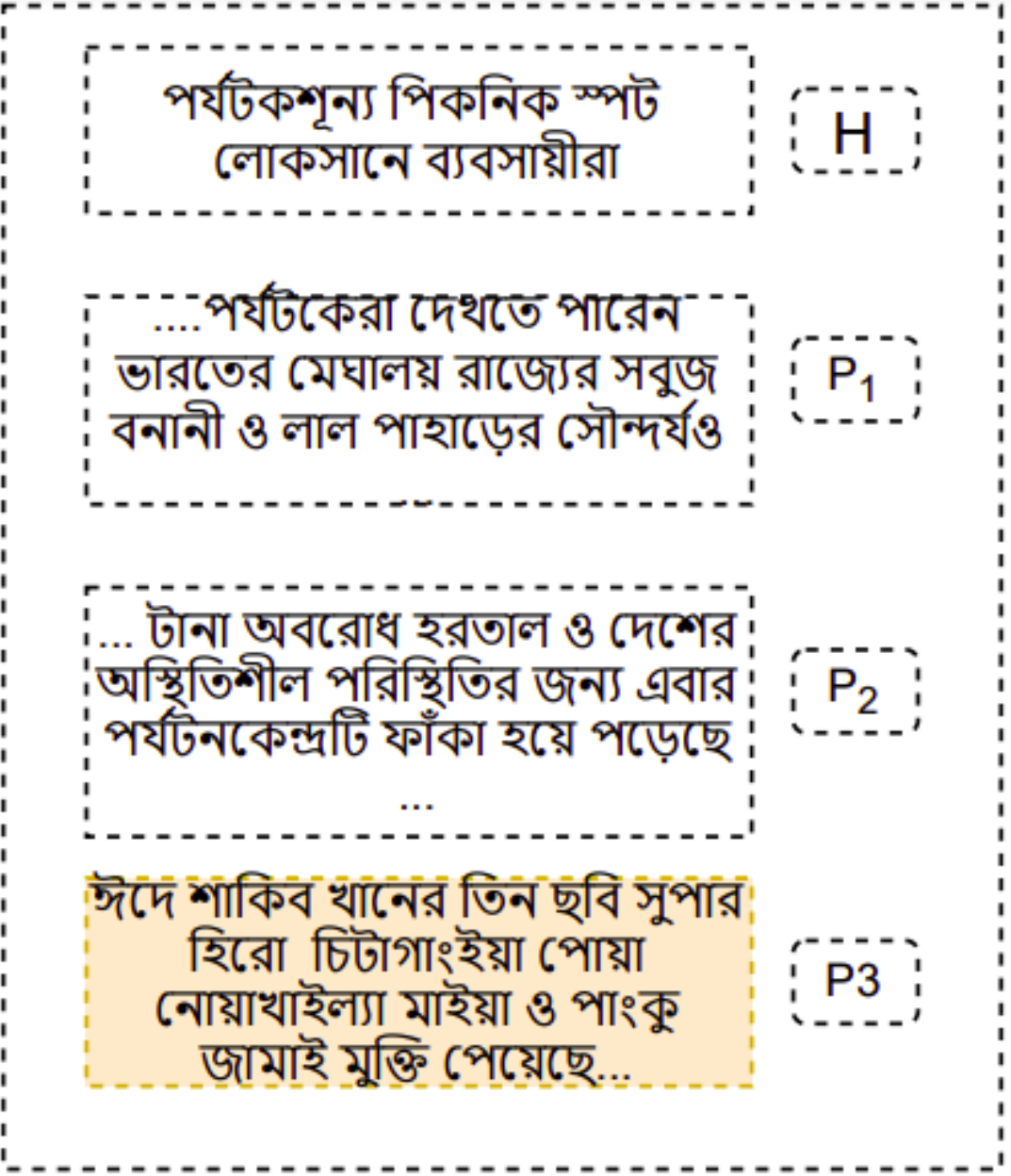}
\vspace{-0.5em} 
\caption{A graph illustration of an incongruent headline problem between the headline and the body paragraphs The inconsistency between paragraph 3 and other texts is represented by a separate color node in the graph.}
\label{into_img}  
\vspace{-1.0em}
\end{figure}

News that is misleading or deceptive has become a major social issue. Much of the information published online is unverifiable, exposing our civilization to unknown dangers. Every day, the amount of news content produced skyrockets. However, unlike newspapers, which only print a certain amount of content each day, making articles online is relatively inexpensive. Additionally, many of these news stories are generated by automated algorithms\cite{doi:10.1177/1461444819858691}, lowering the cost of news production even more. Several news organizations aim to attract readers' focus by employing news headlines unrelated to the main content to draw traffic to news stories among the competition. News headlines are well-known for forming first impressions on readers and, as a result, determining the contagious potential of news stories on social media. People in information-overloaded digital surroundings are less inclined to read or click on the entire content, preferring to read news headlines. As a result, deceptive headlines may contribute to inaccurate views of events and obstruct their distribution. The headline incongruity problem in Bangla news is addressed in this study, which involves determining if news headlines are unrelated to or distinct from the main body text.  Fig.~\ref{into_img} depicts a scenario in which readers could expect to learn precise information about picnic places and picnic spot traders based solely on the headline; however, the news content comprises a Bangla movie advertisement. Because the body text is only accessible after a click, many readers will ignore the discrepancy if they only read the news headlines. Inconsistency in content is becoming more of a concern, lowering the quality of news reading.

Researchers have suggested a number of realistic techniques to address the detection problem as a binary classification utilizing deep learning based on manual annotation (i.e., incongruent or not). A neural network technique is used to learn the features of news headlines and body text in a new way [6]. These techniques, however, face two significant obstacles.For starters, current models focus on recognizing the link between a short headline and a lengthy body text that can be thousands of words long, which makes neural network-based learning difficult.\\

Second, the lack of a large-scale dataset makes training a deep learning model for detecting headline inconsistencies, which involve a variety of factors, more difficult. The headline incongruity problem is solved using a Bangla graph-based hierarchical dual encoder (BGHDE) in this study. It captures the linguistic interaction between a Bangla news headline and any size of body text. By integrating the headline and body paragraph text content as nodes, it makes use of the hierarchical architecture of news stories. On one hand, this method builds a network with nodes for headlines, and on the other hand, it creates a graph with nodes for body paragraphs. Then, between these nodes, we link undirected edges.

The BGHDE is trained to calculate edge weights on a large number of headline and paragraph nodes, with the more relevant edge weight being given. Then, by aggregating information from surrounding nodes, BGHDE updates each node representation. The iterative update technique propagates relevant data from paragraph nodes to the headline node, which is important for spotting content inconsistencies. In this paper we basically followed the data preparation and modeling part from the paper \cite{9363185} and tried to  reproduce the work for Bangla. However, our contribution can be summarized as below:
\begin{enumerate}
  \item We propose a graph-based architecture for the first time to detect incongruity between Bangla’s new headline and body text.
  \item We provide an approach for synthetic data generation and made all the code and pre-processed dataset publicly available\footnote{https://github.com/aminul-palash/bangla-news-incongruity-detection}.
  \item  We have analyzed and presented our model performance both on synthetic dataset and real-world datasets. Apart from news dataset we also tested our model performance, how it can detect irrelevant comments on manually collected comments from social networking cites like Facebook and YouTube.
\end{enumerate}

The findings demonstrate that the proposed method can be utilized to detect inconsistencies in real-world Bangla news reports. As shown in Figure-\ref{into_img}, even for new themes like picnic locations and traders, BGHDE can successfully detect anomalies in headlines and body content. The remainder of this article will be discussed in the following manner: Section 2 begins with a summary of the research on headline incongruity detection and the use of neural graph networks with text. The data creation procedure is then introduced in section 3. Finally, Section 4 discusses the baseline models that were tested in this work. The proposed model is then thoroughly discussed. In Section 5, we give the experimental setup for model evaluation, as well as a discussion of the results obtained from various kinds of Bangla news and empirical study in the field. Finally, we consider the study's implications in terms of combating online infodemics and news fatigue. Finally, Sections 6 and 7 conclude the paper with a discussion of the study's shortcomings as well as future research opportunities in the field of news incongruence detection.

\section{Related Works}

In the present era of the digital world, people are more likely to skim through the headlines and perceive information using both direct memory measures and more indirect reasoning measures\cite{Ecker2014TheEO}. Thus, misleading information can lead to cause more harm. There have been a lot of machine learning algorithms used. Despite the lack of a large-scale realistic dataset being the main issue with this challenge, some studies were able to obtain good results by utilizing manually annotated small-scale datasets. An attention-based hierarchical dual encoder model is used to detect incongruity between the headline and the news body\cite{DBLP:journals/corr/abs-1811-07066}. They have also published a million-scale dataset and introduced augmentation techniques to enhance the training data.

Both incongruity detection and stance detection are related as they share a common basis: to identify the relationship between a brief piece of content and a long piece of content. The goal of the 2019 Fake News Challenge was to encourage the development of stance detection models. For false news identification, a multi-layer perceptron model with one hidden layer\cite{DBLP:journals/corr/RiedelASR17} propagated lexical and similarity features such as bag-of-words (BOW), term frequency (TF), and term frequency-inverse document frequency (TF-IDF). However, the winner of the contest for Fake News Challenge 2019 used the XGBoost\cite{DBLP:conf/kdd/ChenG16} algorithm with extracted hand-crafted features. One of the recent studies used semantic matching\cite{DBLP:conf/icmla/MishraYCL20} dependent on inter-mutual attention by generating synthetic headlines that corresponded to the news body content and original news headline to detect the incongruities. However, for low-resource languages like Bengali, this kind of work is rare due to a lack of well-developed datasets. In the paper\cite{DBLP:conf/lrec/HossainRIK20} , they explored neural network models and pre-trained transformer models to detect fake news for fake news detection in Bangladesh. They've also released an annotated Bangla news dataset that can be used to create an automated fake news detector.

Graph neural network (GNN) is semi-supervised learning which utilizes graph-structured data \cite{DBLP:journals/corr/KipfW16}. GNN models can learn hidden layer representations by encoding local graphs and features of vertices while maintaining the linear model scales of graph edges. The fundamental advantage of GNN models over traditional models such as recurrent neural networks (RNN) and convolutional neural networks (CNN) is that GNN models embed relational information and pass it on to neighbor nodes during training. As a result, GNN models have succeeded miraculously in NLP tasks like question-answering
\cite{DBLP:journals/corr/abs-1809-02040, DBLP:journals/corr/abs-1905-06933} , relationship extraction \cite{DBLP:conf/emnlp/Zhang0M18} and knowledge base completion \cite{DBLP:conf/esws/SchlichtkrullKB18}.

Fake news detection using GNN models has become a common practice nowadays. The paper \cite{DBLP:conf/icde/ZhangDY20} introduced FAKEDETECTOR, an automatic fake news detector model based on explicit and hidden text features that constructs a deep diffusive network model to simultaneously learn the representations of news articles, producers, and subjects.The Hierarchical Graph Attention Network was also used by the researchers, which use a hierarchical attention method to identify node representation learning and then employs a classifier to detect bogus news. \cite{DBLP:conf/ijcnn/RenZ21}.

To the best of our knowledge, our work is the first to detect incongruity between news headline and body for Bangla. In our work, we proposed a Bangla graph-based hierarchical dual encoder (BGHDE) model for automatic detection of Bangla news headline and body. 

\section{Datasets for detecting Headline In-congruence Problem}

We propose an approach for detecting incongruity between Bangla news headlines and content where we specifically tackle three significant challenges for preparing the dataset. We followed the same approach from this paper \cite{9363185} for preparing our dataset for bangla language. The first is the scarcity of manually annotated training datasets, as well as the high expense of creating them. The second is the length of news stories, which can often be long and arbitrary, making them challenging to model for machine learning. The last one is the paragraph creation from the Bangla news corpus, as our news dataset doesn’t contain any paragraph separation. In the sections below, we'll go over each obstacle in detail.

There are millions of news articles over the internet, and previously, the ground truth was manually annotated in earlier investigations \cite{inproceedings}, \cite{inproceedingsdataset}. However, it is almost impossible to annotate ground truth for each news manually. Therefore, although some previous studies created manually annotated datasets, we adopt an automated approach for creating the annotations.

To begin with, we gather news stories from reputable target news sources. Then we pick a collection of target news articles at random to manipulate, i.e., change them to look like incongruent news sources.

For the news pieces chosen for alteration, we replace some paragraphs with paragraphs obtained from other news sources. This collection of parts is created individually and is not used for training or testing. We carefully monitor the alteration process to ensure that no news stories are duplicated.


\begin{figure}[H]
\centering
\includegraphics[width=1\linewidth]{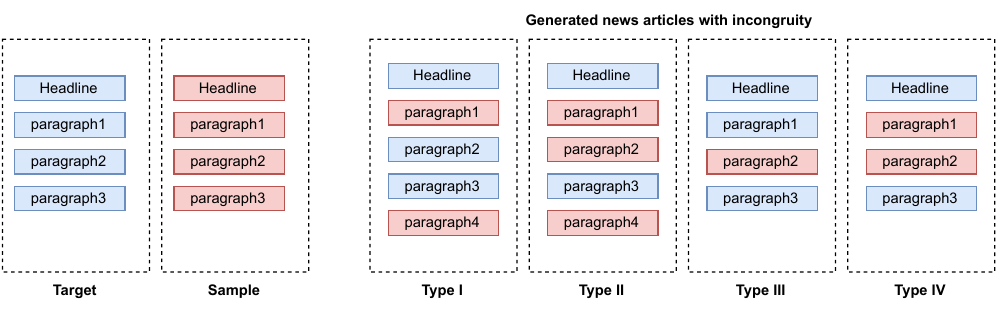}
\vspace{-0.5em}
\caption{An illustration of how news items with incongruent headlines are created. Paragraphs from samples articles mixed with target news articles randomly. Total five types of data mix-up processes shown (Type I, Type II, Type III, and Type IV).}
\label{fig:data-gen}  
\vspace{-1.0em}
\end{figure}

The news corpus that we use in this paper comes from a renowned 
The news corpus that we employ in this paper comes from the famous Bangla news site Prothom Alo\footnote{https://www.kaggle.com/furcifer/bangla-newspaper-dataset}, which consists of over 400k authentic Bengali news articles.

The figure \ref{fig:data-gen} shows The overall workflow of incongruent level data preparation \cite{9363185}. There is one target news article and one sample news article. In our case, we select sample news articles randomly from the real news corpus. The sample news content paragraphs are then blended in with the intended news content. The maximum number of paragraphs that can be switched is determined by the number of paragraphs in the target news article. We randomly manipulate paragraph swapping processes that address different difficulty levels on the target news article.

 From Figure \ref{fig:data-gen}, there is a total of four types of samples that can be generated by the generation process. But we don’t pick the first two types (I and II) for preparing our training datasets so that we can keep consistent same distribution with the original and generated news content in length. So we keep types three and four (III and IV) for preparing our synthetic training datasets.

\begin{table}[H]
\centering
\setlength{\tabcolsep}{10pt}
\caption{Statistics of our Bangla data}
\scalebox{1}{

\begin{tabular}{ll|cc|cc}

\toprule
&  & \multicolumn{2}{c|}{\textbf{Headline}} & \multicolumn{2}{c}{\textbf{Content}} \\
\midrule
\textbf{Dataset}                & \textbf{Samples}  & \textbf{Avg.} &  \textbf{Std} & \textbf{Avg.} &  \textbf{Std}   \\ 
\midrule
Train   & 228000  & 5.58 & 1.45 & 319.35 & 205.41 \\
Dev   & 120000  & 5.58 & 1.43 & 319.01 & 241.06 \\
Test   & 120000  & 5.57 & 1.43 & 323.55 & 214.124 \\
\bottomrule
\end{tabular}
}
\label{tab-dataset}
\end{table}

As there was no paragraph separation on our collected dataset, we had to separate paragraphs synthetically based on the number of sentences on the news content. Thus, we split the news article by paragraph, where each paragraph contains five, ten, twenty, and so on sentences based on the length of news content.

Furthermore, we preprocessed the data by removing unnecessary punctuation from both headline and body content. We also discarded the news data that contains small body content.

\section{Methodology}
Our goal is to determine if the content of the news article matches the news headline. First of all we extracted semantic information from the data as the distributed representation of words and sub-word tags has proven effective for text classification problems. Therefore,
we used pre-trained bangl word embedding, where an article is represented by the mean and the standard deviation of the word vector representation. For this experiment we used Bengali GloVe 300-dimensional word vectors pre-trained embedding\footnote{https://github.com/sagorbrur/GloVe-Bengali}, and our coverage rate was 43.34\%. 

For detecting news incongruities, we adopt a Graph Neural Network (GNN) based architecture following \cite{9363185}. GNN is a type of deep learning method that can be used for processing data represented as a graph \cite{scarselli2008graph} for making node-level, edge-level, and graph-level analyses.

\begin{figure}[H]
\centering
\includegraphics[width=1\linewidth]{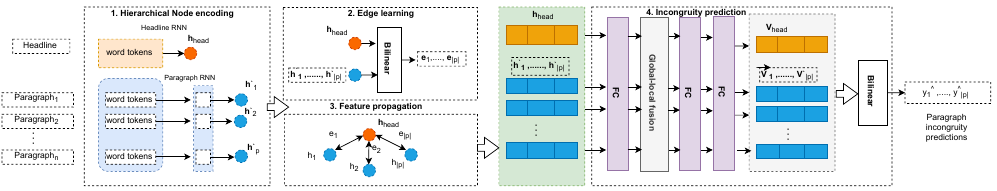}
\vspace{-0.5em}
\caption{An overview of our proposed model for Bangla news incongruity detection.}
\label{fig:model_arch}  
\vspace{-1.0em}
\end{figure}

Our proposed Bangla graph-based hierarchical dual encoder (BGHDE) illustrated in figure \ref{fig:model_arch} takes the headline and paragraph contents into account to detect incongruency in an end-to-end manner. Our proposed architecture is followed from this paper \cite{9363185} which implement an incongruity detection method for english dataset.The architecture consists broadly of four steps which we describe briefly below.

First of all, the BGHDE creates a graph that is undirected representing its innate structure for each news article, After that, it was utilized to train the neural network with graph structure. Finally, a hierarchical Gated Recurrent Unit (GRU)-based bidirectional Recurrent Neural Network(RNN) framework is utilized to generate a representation with node structure of each headline text and paragraph text, as well as context-aware paragraph representation.

A group of nodes called vertex represents the headlines text and each corresponding paragraph of the Bangla news material. The edge of the graph represents the link between headlines and its corresponding paragraphs of the Bangla news.

Secondly, to avoid undesirable flattening of the node representation between the congruent and incongruent paragraphs during GNN propagation, the next step is to learn the edge weights of the input graph representation.

We employ the paragraph congruity value as a label to supervise the edge weights during the cross-entropy loss.

Because of this edge-level monitoring, the BGHDE can assign larger weights to congruent paragraphs and lower weights to incongruent paragraphs, allowing congruent paragraphs to communicate more information to the headline node than incongruent paragraphs alone.

Later, the node characteristics are transmitted into surrounding nodes in the third phase using the established graph structure and trainable edge weights from the GNN framework. In an edge-weighted form, BGHDE uses the convolutional graph network (GCN) aggregation function. \cite{kipf2017semisupervised}

Finally, the incongruity of scores of news pieces is predicted for the last stage. The GNN graph classification problem is the same as this. The global-level graph representation and the local-level node representation must be fused together, BGHDE adapts a fusion block presented in\cite{wang2019dynamic}

\section{Experiments}
\begin{figure}[H]
\centering
\includegraphics[width=0.85\linewidth]{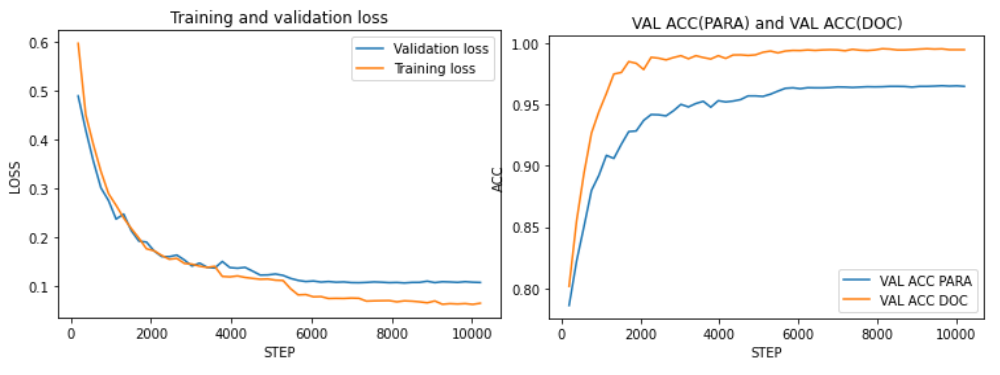}
\vspace{-0.5em}
\caption{Left plot represents training and validation loss and the right plot represents accuracy of the validation data during training.}
\label{fig:model-loss}  
\vspace{-1.0em}
\end{figure}

The title and the appropriate paragraphs of the subsequent body text are encoded using a single-layer GRU with 200 hidden units. In contrast, a single-layer bidirectional GRU with 100 hidden units is used to encode the paragraph-level RNN. There are three GNN layers, each with 200 hidden units. Hidden unit dimensions of 200, 200, and 100 for the FC layers applied after feature propagation on the graphs, respectively. We use the Adam optimization algorithm \cite{kingma2014adam} to train the model, starting with an initial learning rate of 0.001, which is decayed every three epochs by a factor of 10. We use 120 samples for each mini-batch during training. We clip the gradients with a threshold of 1.0. The hyperparameter for tradeoffs for edge loss is set to 0.1. We use pre-trained Bangla\footnote{https://github.com/sagorbrur/GloVe-Bengali} GloVe \cite{pennington2014glove} embedding consisting of 300-dimensional vectors to initialize the word embeddings.The number of words that appear at least eight times in the training dataset determines the size of the embedding matrix's vocabulary. The model has 1,214,702 total trainable parameters. We obtained an accuracy of 0.9560 and an AUC score of 0.9860 on the validation set. We show the loss and paragraph and document accuracy obtained during training in Figure \ref{fig:model-loss}.

We use PyTorch \cite{Paszke2019PyTorchAI} and PyTorch Geometric \cite{Fey2019FastGR} frameworks to implement the model and Google colab to run the experiments.





\section{Results and Discussion}

To validate our process, we conduct rigorous quantitative and qualitative analyses. We conducted a large-scale experimental analysis to evaluate our proposed Bangla graph-based hierarchical dual encoder (BGHDE). 
We performed an evaluation based on criteria like accuracy on both paragraph-wise and whole document or news article calculated while all types of a common evaluation matrix for the predictive model also added.

\begin{figure}[H]
\centering
\includegraphics[width=1\linewidth]{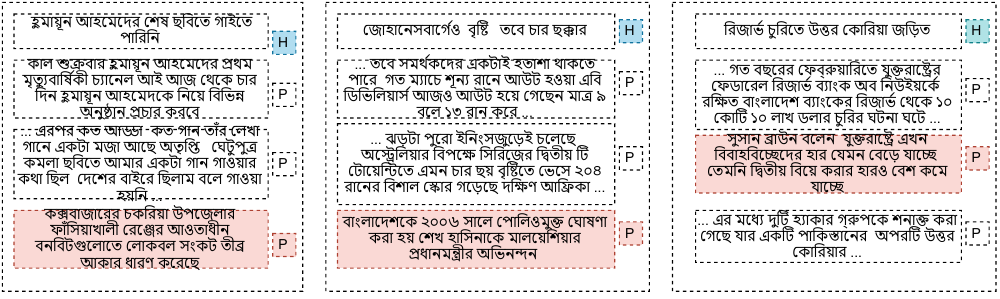}
\vspace{-0.5em}
\caption{Examples of some partial representation of news articles that contains incongruity between headline and content paragraphs, colored paragraph are the in-congruence paragraph}
\label{fig:data-examples}  
\vspace{-1.0em}
\end{figure}

\begin{figure}[H]
\centering
\includegraphics[width=1\linewidth]{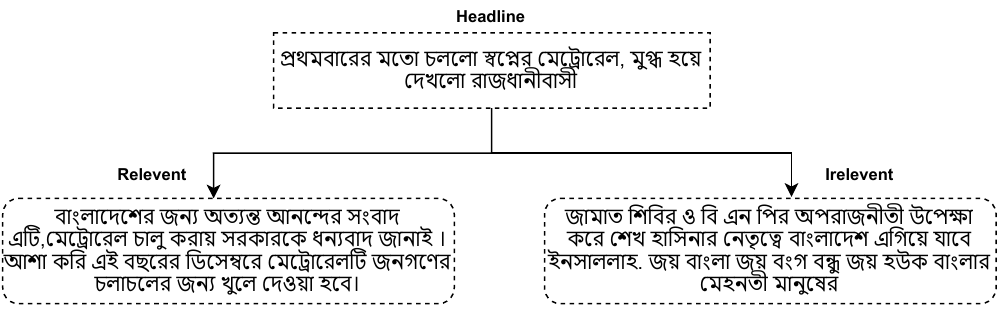}
\vspace{-0.5em}
\caption{Incongruity detection on comments data}
\label{fig:data-comment}  
\vspace{-1.0em}
\end{figure}

\subsection{Performance Evaluation on Synthetic Dataset}
Figure \ref{fig:data-examples} illustrates some examples of incongruent data detected by our proposed model on synthetic test datasets. Our tested dataset contains news from both Bangladeshi\footnote{https://www.kaggle.com/ebiswas/bangla-largest-newspaper-dataset} and West Bengal\footnote{https://www.kaggle.com/csoham/classification-bengali-news-articles-indicnlp} news articles which helps to analyze the model performs better Table \ref{tab-results} shows the model’s performance on different datasets when it was used to detect headline inconsistencies.

\begin{table}[ht]
\setlength{\tabcolsep}{4pt}
\centering
\caption{Second table}
\scalebox{1}{

\begin{tabular}{lccc|cccc} 
\toprule
\multirow{2}{*}{Dataset} & \multirow{2}{*}{Size}  & \multirow{2}{*}{Acc(para)}  & \multirow{2}{*}{Acc(doc)}  & \multicolumn{4}{c}{\textbf{Evaluation}} \\
& & & & Precision & Recall & F1 Scr & Support \\
\midrule
Prothom alo & 6000 & 0.9658 & 0.9918 & 0.98.80 & 0.9956 & 0.9918 & [3000 3000]  \\
bdnews24 & 2000 & 0.9175 & 0.94.50 & 0.7554 & 0.913 & 0.9431 & [1000 1000]  \\
Ananda bazar & 1000 & 0.9175 & 0.9450 & 0.9623 & 0.97 & 0.9431 & [500 500]  \\
  
ebela & 5000 & 0.9192  & 0.9702 & 0.970 & 0.9704 & 0.9702 & [2500 2500]   \\
zeenews & 5000 & 0.9026 & 0.9542 & 0.9511 & 0.9576 & 0.9543 & [2500 2500] \\
Ittefaq & 8000 & 0.9445 & 0.9866 & 0.9812 & 0.9922 & 0.9866 & [4000 4000] \\
Jugantor & 6999 & 0.9494 & 0.9862 & 0.9830 & 0.9893 & 0.9861 & [3458 3477]  \\

\bottomrule
\end{tabular}
}
\label{tab-results}
\end{table}

\subsection{Evaluation on Real world Dataset}
To see how effective our dataset and proposed models are in detecting incongruent headlines in the real world. We’ve conducted this process by collecting data containing actual articles in which any form of the generation process has not modified the body text. 
It is difficult to annotate read news containing incongruity manually, and we perform inference on a real news dataset without annotations. After that, we evaluated our model performance manually. But we achieved very poor performance. For example, the datasets we used have no paragraph separation, and we need to separate the articles into paragraphs randomly. Another one is that we only train our model on synthetic datasets.

We also evaluated our proposed model on detecting incongruent comments Figure \ref{fig:data-comment} on different sites like YouTube, Facebook, and various Bangla news sites. As a result, we collected more than two hundred comments data containing relevant and irrelevant comments with corresponding news articles.
We achieved an accuracy of 0.73 on the Bangla comments dataset.

\section{Conclusion and Future Work}
For the first time, a graph neural network was used to handle the headline incongruity problem in Bangla news stories. We discovered a few false-positive scenarios when a model misinterpreted a coherent article for an incongruent headline using manual annotations. Although the computer accurately predicted the label, according to the idea of headline incongruity, such a "briefing" item does not mislead readers by delivering false information. Our findings suggest that more research is needed in the future to improve data generation and gathering processes. The findings of the evaluation experiments, however, reveal that the proposed technique accurately detects such deception. We hope that our research helps to create more trustworthy online news ecosystems in Bangla.

\bibliographystyle{502}
\bibliography{502}
%





\end{document}